\documentclass[conference]{IEEEtran}
\ifCLASSOPTIONcompsoc
  \usepackage[nocompress]{cite}
\else
  \usepackage{cite}
\fi
\ifCLASSINFOpdf
\else
\fi
\usepackage{amsmath}
\usepackage[utf8]{inputenc}
\usepackage[LAE,T1]{fontenc}

\usepackage[flushleft]{threeparttable}

\usepackage{booktabs}% http://ctan.org/pkg/booktabs

\usepackage{graphicx}
\usepackage{ragged2e}
\usepackage{makecell}
\usepackage{multirow}
\usepackage{array, tabularx}
\newcommand{\arabiclr}[1]{\bgroup\beginR\fontencoding{LAE}\fontsize{0.3cm}{0.3cm}\selectfont#1\endR\egroup}

\newcommand{\arabicrl}[1]{\bgroup\beginR\raggedleft\fontencoding{LAE}\fontsize{0.2cm}{0.3cm}\selectfont#1\endR\par\egroup}

\makeatletter
\let\old@ps@headings\ps@headings
\let\old@ps@IEEEtitlepagestyle\ps@IEEEtitlepagestyle
\def\confheader#1{%
	\def\ps@IEEEtitlepagestyle{%
		\old@ps@IEEEtitlepagestyle%
		\def\@oddhead{\strut\hfill#1\hfill\strut}%
		\def\@evenhead{\strut\hfill#1\hfill\strut}%
	}%
	\ps@headings%
}
\makeatother

\confheader{%
	{\small 11th International Conference on Computer and Knowledge Engineering (ICCKE 2021), October 28-29, 2021, Ferdowsi University of Mashhad}
}

\usepackage[pscoord]{eso-pic}
\newcommand{\placetextbox}[3]{
	\setbox0=\hbox{#3}
	\AddToShipoutPictureFG*{ \put(\LenToUnit{#1\paperwidth},\LenToUnit{#2\paperheight}){\vtop{{\null}\makebox[0pt][c]{#3}}}
	}
}
\placetextbox{.23}{0.055}{\small{978-1-6654-0208-8/21/\$31.00 ~\copyright2021 IEEE}}
\begin{document}
\title{ExaASC: A General Target-Based Stance Detection Corpus in Arabic Language}

\author{\IEEEauthorblockN{Mohammad Mehdi Jaziriyan}
	\IEEEauthorblockA{Exa Company\\
		Tehran, Iran\\
		Email: m.jaziriyan@exalab.co\\
		Email: m\_jaziryan@modares.ac.ir}
	\and
	\IEEEauthorblockN{Ahmad Akbari}
	\IEEEauthorblockA{Exa Company\\
		Tehran, Iran\\
		Email: a.akbari@exalab.co\\
		Email: ahmadakbari@ut.ac.ir}
\and
\IEEEauthorblockN{Hamed Karbasi}
\IEEEauthorblockA{Exa Company\\
	Tehran, Iran\\
	Email: karbasi@exalab.co}
    Email: karbasi\_hamed@ee.sharif.edu}


% compsoc conferences compared to traditional conferences), use this
%Telephone: (800) 555--1212, Fax: (888) 555--1212}


\maketitle

\begin{abstract}
Target-based Stance Detection is the task of finding a stance toward a target. Twitter is one of the primary sources of political discussions in social media and one of the best resources to analyze Stance toward entities. This work proposes a new method toward Target-based Stance detection by using the stance of replies toward a most important and arguing target in source tweet. This target is detected with respect to the source tweet itself and not limited to a set of pre-defined targets which is the usual approach of the current state-of-the-art methods. Our proposed new attitude resulted in a new corpus called ExaASC for the Arabic Language, one of the low resource languages in this field. In the end, we used BERT to evaluate our corpus and reached a 70.69 Macro F-score. This shows that our data and model can work in a general Target-base Stance Detection system.
The corpus is publicly available\footnote{https://github.com/exaco/exaasc}.

\end{abstract}


% \fi
\begin{IEEEkeywords}
Target-based Stance Detection, Stance Detection, Arabic, Corpus, BERT

\end{IEEEkeywords}

\IEEEpeerreviewmaketitle

\section{Introduction}\label{introduction}
Social media are an essential source of opinions, information, and ideas. People talk about politics, sports, personal activities in their daily lives, and Twitter is one of the primary resources to express views and opinions. In Target-based Stance detection, the Stance is expressed towards one or more targets \cite{mkszc16}. Targets can be different types of entities like a person, event, and claim. Rumor-based Stance detection uses tweets as a claim and stance of replies determined toward the claim \cite{dblphz17}, but in this paper, we propose a new approach for target-based stance detection by using Twitter source tweets and their replies. We choose the most important target in the source tweet, then uses its replies for detecting Stance towards it. Because of limited contextual information in a short text, e.g., tweet, we can gain necessary information about the target in the source tweet and use replies to detect its target's Stance.

Until now, there are no public datasets for target-based Stance detection in the Arabic Language, so we provide a new corpus for target-based stance detection (ExaASC) in this language that contains different types of targets like persons, entities, and events.

In recent years, with the introduction of Pre-trained Language Models, Natural Language Processing enters a new era. Language Models such as BERT \cite{dclt19}, GPT\cite{radford2018improving}, ELMo\cite{peters-etal-2018-deep}, etc., trained on large text corpora to enhance the semantic representations and context relationship between words of sentences and improved accuracy of downstream tasks in the Natural Language Understanding domain that shifts the quality of its tasks to a better position. This research has fine-tuned the Arabic BERT model to detect the stance of replies toward targets annotated in main tweets.

A major difference in this research with others is that we use one model for a wide variety of targets instead of one model for each target. Thus, it has a better generalization ability, but this results in a lower F-score than the latter.

We used instructions in paper \cite{mkszc16slovenia} for annotating stance labels for our dataset with a minor difference, Favor and Against labels tagged based on paper but None class contains tweets without explicit or implicit stances (\textit{Neutral}); moreover, irrelevant tweets (\textit{Neither}).

Overall, the main contributions of our work are:
\begin{itemize}

\item Provide ExaASC a new general corpus for Arabic target-based stance detection with different types of targets
\item New Target-base attitude for Stance detection using tweets and their corresponding replies
\item Comparing different kinds of pre-trained and deep learning-based models on the dataset.
\item Fine-tuning Arabic BERT for the stance detection task.
\end{itemize}
The rest of the paper is divided as follows. In the following section, first, we discuss related works in stance detection. In section \ref{dataset prepration}, we describe our dataset, dataset annotation, and results. In section \ref{model}, we explain the model and used architectures and in section \ref{experimental setups} we describe model configuration and hyperparameters used in this paper. In section \ref{experimental results} we discuss our experimental results, and finally, in the last section, we conclude our work and provide future research directions.

\section{Related Works}
In this section, we provide related works that worked on stance detection. Older research on stance detection used debates in online forums using traditional feature extraction methods and machine learning algorithms. In recent years researchers primarily focus on representation learning methods such as neural networks and modern pre-trained language models for the stance detection task.

In the SemEval-2016 Stance Detection task, Mohammad \textit{\textit{et al.}} \cite{mkszc16} used a Target-based stance detection dataset with sentiment tags that proposed in \cite{mkszc16slovenia} for two subtasks: subtask A was for supervised stance detection and consisted of 5 target topics (legalization of abortion, feminism, Hillary Clinton, Climate change, Atheism) 4163 tweets overall for SemEval competition. Subtask B was for weakly supervised stance detection, providing extensive unlabeled data about 78000 tweets and a small test set with 700 tweets for another target (Donald Trump). Competitors propose several supervised and weakly supervised methods for this competition and the best score paper for subtask A was the paper that used two-staged RNN first trained on large Twitter corpus and second initialized by first RNN and trained on provided data and attains F-score 67.81\% \cite{zm16}. For subtask B, Wei \textit{et al.} \cite{wzlcw16} used Convolutional Neural Network for text and achieved F-score 56.28\%. The baseline method for subtask A was an SVM-based method with n-gram features that had a 68.98\% F-score.

Sobhani \textit{et al.} \cite{smk16} proposed an SVM classifier with features of the word and character n-grams as well as sentiment features like sentiment lexicons and Word2Vec embeddings. Their method outperforms all teams that participated in the SemEval-2016 task and got a Micro-F-score of 75.3\%.

The proposed model in \cite{msk17} used sentiment tags for improving stance detection and improved 5 percent on SemEvel2016. This model has better accuracy on texts which have different topics.

Another dataset on this topic created a multi-Target dataset in which one text has multiple Topics. The data was for the US election 2016, and topics were about the US candidates. The proposed model in \cite{siz17} used a seq2seq model, which encodes tweets in an RNN and decodes it in an ANN model. The output is a sequence of tags for each topic in the text. They’ve reached an F-score of 54.8\%.

One of the SemEval-2016 participants who received 9th rank \cite{khs17}, produced a new dataset on Czech Language and tested their proposed model in that Language. This dataset has 1550 news texts from the Czech with two targets. The Kappa agreement of this dataset was 0.66. They used n-gram and \textit{tf-idf} features in their model, and they used 1000 most frequent words and their sentiment tag in training. Their final accuracy on two topics was (0.43, 0.46) which was too low. 

SemEval-2017 stance detection task provides a rumor stance detection dataset in which the target is the whole tweet text and tags stance for reply corresponding to it. It has 4519 tweets \cite{dblphz17}.  Two years later, RumourEval-2019 added Reddit discussions and released 9000 samples like the previous RumourEval event \cite{gklazbd19}.

The authors of the paper \cite{lcfbpr20} created a multilingual dataset consisting of English, Spanish, French and Italian Languages. For each language, they had one or two targets, and the whole dataset had 14440 texts. They used BiLSTM, CNN, LSTM, LR, and SVM methods and got an accuracy of between 0.45 to 0.68 on different topics and languages.

There is a multi-task Attention-based model proposed in \cite{lc19}, which trains Attention besides feeding important sentiment words and applying this model for train and use on Stance-based Target. Another architecture proposed in \cite{wslzz20} is a hierarchical model that contains Hierarchical Attention layers and one Attention layer above all of them. This model uses and weights outputs of NLP modules such as sentiment analysis, dependency parsing, and agreement in the model.

There is one dataset in the Arabic Language for Target-based stance detection proposed by Darwish \textit{et al.} \cite{dmz17}. The authors chose the controversial issue of transferring ownership of the Tiran and Sanafir islands from Egypt to Saudi Arabia, including about 33,000 tweets for 2,400 users. This dataset has only one target.

Ghosh \textit{et al.} \cite{gssrg19} analyzed several methods and compared them. These methods are CNN, Target-Specific Attention Neural Network (TAN), LSTM, SVM, Two-Step SVM, and BERT.  It's best macro average of the F-score on SemEval2016 was for the BERT model, and it was 0.751. The second score of these methods was for two-step SVM, which was 0.744.

Eventually, a summary of the mentioned datasets, number of tweets, number of targets and their languages is shown in Table \ref{related work table}.

\begin{table}[thb]
	\renewcommand{\arraystretch}{1.5}
	\caption{Some of the datasets in Stance Detection}
	\label{related work table}
	\centering
	\begin{tabular}{|l|c|c|c|}
\hline
\textbf{Dataset}        & \textbf{\thead{Number of\\ docs}} & \textbf{\thead{Number of \\targets}} & \textbf{Language} \\ \hline
\textit{SemEval-2016\cite{mkszc16}}   & 4870  & 6                   & English           \\ \hline
\textit{Island Dataset\cite{dmz17}} & 33024        & 1                   & Arabic            \\ \hline
\textit{News Comments\cite{khs17}} & 1560        & 2                   & Czechc            \\ \hline
\textit{RumorEval-2017\cite{dblphz17}} & 4519         & Various Rumors      & English           \\ \hline
\textit{RumorEval-2019\cite{gklazbd19}} & 9000         & Various Rumor       & English           \\ \hline
\textit{\thead[l]{Multi-lingual\\ dataset\cite{lcfbpr20}}} & 14440         & \thead{1 or 2 per \\each language}       & Multi-lingual \\ \hline
	\end{tabular}
\end{table}

\section{Dataset Prepration}\label{dataset prepration}
\subsection{Data Annotation Guide}
Tweets and the replies were extracted from Twitter stream API. From about 80,000 source tweets, the controversial and mostly argumentative tweets were selected. The target of the source sentence is determined by our professional annotator based on what replies talk about in the source sentence. Between multiple targets in a tweet, the most related one is selected. The Stance tags were annotated using Mohammad \textit{et al.} paper \cite{dclt19} guidelines. Summary of annotated labels are as follows: 
\begin{itemize}
\item \textit{Favor} label if reply explicitly or implicitly supports the target or supports an entity related to the target.
\item \textit{Against} label annotated if the reply directly was against the target or indirectly against the entity that relates to the target.
\item \textit{None} label if Stance toward target does not express or does not relate to the target.
\end{itemize}

Some samples of the proposed dataset with English translation have been exhibited in Table \ref{dataset sample table}.
%% Dataset sample table

\begin{table*}[!htb]
	\renewcommand{\arraystretch}{1.5}
	\caption{ExaASC Samples}
	\label{dataset sample table}
	\centering
	\begin{threeparttable}
		\centering
		\begin{tabularx}{\textwidth}{|X|X|c|c|}
			\hline
			\textbf{Source} & \textbf{Reply} & \textbf{target} & \textbf{Stance Tag} \\
			\hline
			\arabicrl{
					توني كروس : كأس العالم في قطر قرارًا خاطئًا..
					العمال يعملون بلا توقف ، درجة الحرارة 05 ، مع سوء التغذية وعدم وجود ماء !!!!
			}
		Tony Kroos: The World Cup in Qatar is a wrong decision. The workers are working non-stop, the temperature is 50 degrees in addition to malnutrition and lack of water.
		 & 
			\arabicrl{مااتوقع لاعب بحجم كروس يقول هالكلام بالعكس عنده خلفيه جميله عن العمل القائم بقطر 
					المصدر لو سمحت ولا حاب البرشلونيه يسون حمله على كروس كلام فاضي!! 
				}
			
			I do not think a player the size of Kroos would say such a thing. On the contrary, he has a good background to the moves made in Qatar. Please tell the source of the news. I do not like the Barcelona fans to attack Kroos
			
			 & \thead{\arabicrl{توني كروس} \\  
			 Tony Kroos}
			 
			 & Favor \\ 
			%\hline
			\cline{2-4} 
			& \arabicrl{والله ياكروس خوينا صاير غبي مالك دخل} 
			Kroos swears to God our brother is stupid. Don't worry
			 &  \thead{\arabicrl{توني كروس} \\  
			 	Tony Kroos}
			 	& Against \\
			\cline{2-4}
			& \arabicrl{كلامه سليم 001٪} 
			His words are 100\% true
			 &  \thead{\arabicrl{توني كروس} \\  
			 	Tony Kroos} & Favor \\
		\cline{2-4}
      & \arabicrl{السلام عليكم ورحمة الله متوفر افضل الاسعار لتجديد اشتراكات بي ان ،حياكم الله عبر الخاص او عن طريق الواتس اب. خدماتنا ليست محصورة على تجديد الاشتراكات فقط بل يتم خدمتكم بعدة خدمات بعد التجديد}
      Hello ... The best prices for re-subscribing in BN .. Welcome to Direct or WhatsApp ... Our services for re-subscription are not small, only after subscribing you will be given more services
  
			 & \thead{\arabicrl{توني كروس} \\  
			 	Tony Kroos} & None \\
     \hline
      \arabicrl{عاجل  . . وزير الاتصالات : المملكة العربية السعودية اليوم في المرتبة الثانية عالمياً في تقنية 5 جي. . . السعودية }
breaking news . . Minister of Communications: The Kingdom of Saudi Arabia is today ranked second in the world in 5G technology. . . Saudi Arabia
    & \arabicrl{ماشاء الله لا قوة الا بالله}
    	Masha Allah, there is no power but with Allah
    & \thead{\arabicrl{وزير الاتصالات} \\ 
    Minister of Communications }
    & Favor \\
      \cline{2-4}
%      & \arabicrl{قولوا له تعال قرى ومحافظات الجنوب مثل قرى محافظة القنفذة وتحديدا في القوز وما جاورها وتجدنا مازلنا في تقنية gprs اي التقنية في اول بدايات دخول النت للمملكة اي ما قبل G1الله يوفقكم ويوفقه والتي اذا طفشت النت منك اعطتك اخر قوتها بالرمز E قرى القنفذة تحتاج خدمة جيدة في النت ً
%      }
%  Tell him, “Come to the villages and governorates of the south, such as the villages of Al-Qunfudhah Governorate, specifically in Al-Quoz and its environs, and you will find us still in the technology of gprs, which is the technology in the first beginnings of the Internet’s entry into the Kingdom.That is, before G1, may God bless you and grant him success, which if your internet becomes overburdened, will give you the last of its strength with the symbol E. Qunfudhah villages need good internet service 
%  & \thead{\arabicrl{وزير الاتصالات} \\ 
%  	Minister of Communications }
%  & Against \\
%  \cline{2-4}
  & \arabicrl{وش ذا الكذب يا معالي الوزير
  }
What is this lie, Mr. Minister?
   & \thead{\arabicrl{وزير الاتصالات} \\ 
   	Minister of Communications } 
 & Against \\
  \cline{2-4}
  & \arabicrl{ويلك والله باكستان احسن منا مدري لمتى تكذبون الكذبه وتصدقونها}
  Oh God, even Pakistan is better than us, how long do you want to lie and believe it 
     & \thead{\arabicrl{وزير الاتصالات} \\ 
     	Minister of Communications }
  & Against \\
  \hline
		\end{tabularx}
	\end{threeparttable}
\end{table*}
\subsection{Dataset Statistics}
We select 100 stance data samples verified on our at least three professional annotators for the target-based Stance detection dataset annotations. We use this data to take a test from candidates and select top-performing annotators. We also had multiple sessions to check annotators tag's correctness, fix their mistakes, and get everyone on the same page. Overall data consist of about 17500 tagged data. Still, we select 9566 samples tagged by at least two annotators and put aside non-agreement samples and one tag sample for improving the data quality. This dataset was annotated by at least two native Arabic annotators trained on Stance detection annotation guidelines.  Training data consisted of 6826 samples with about more than 180 unique targets (test set contains 20 targets of aforementioned), targets mainly are about political persons, events, and a few sports targets. Data consists of 360 source tweets with their replies. 10\% of this training data is used for validation set and does not have overlap with source tweets in training data. The remaining 2740 data used for the test set and also do not have overlap. 20\% of data dropped because of disagreement between annotators. We tried to balance our data on different classes, but because of the nature of the None class, this class consists of most of the data (30\%). 

The data format is as follows: \textit{id, main, reply, target, annotators\_id, majority\_tag}. We repeat the source tweet(\textit{main}) to the number of corresponding replies to optimize the model efficiently. \textit{annotator\_id} consists of the tags of each annotator and its corresponding id in the system. \textit{majority\_tag} is the aggregation of annotators tags.

The system used for tagging dataset is the utag\footnote{https://utag.ir} website developed by Exa corporation to tag and annotate data samples. 
The statistics of the dataset are presented in Table \ref{dataset stats}.
\begin{table}[thb]
	\renewcommand{\arraystretch}{1.5}
	\caption{ExaASC Dataset Stats}
	\label{dataset stats}
	\centering
	\begin{tabular}{|l|c|c|c|c|}
		\hline
		\bfseries Data & \bfseries Favor & \bfseries Against & \bfseries None & \bfseries All 
		\\
		\hline
		\bfseries Train & 2628 & 1898 & 2300 & 6826
		\\
		\hline
		\bfseries Test & 682 & 1012 & 1046 & 2740
		\\
		\hline
		\bfseries Overall & 3310 & 2910 & 3346 & 9566
		\\
		\hline
	\end{tabular}
\end{table}

\section{Model}\label{model}
In this section, we first discuss the BERT model, then we describe architectures on top of BERT that we used for applying our data.
\subsection{BERT}
BERT \cite{dclt19} is a Bidirectional Transformer Encoder trained on massive unlabeled text corpora with two training objectives: Masked Language Model (MLM) and Next Sentence Prediction (NSP). For the MLM objective, BERT masks 15\% of tokens to predict the id of masked vocabs. For the NSP objective, BERT uses consecutive sentences 50 percent of the time and random sentences for other parts, predicting whether two sentences are consecutive or not. In recent years, BERT has been utilized as a pre-trained Language model for Downstream tasks like sentiment prediction, question answering, language inference, etc. For Fine-tuning BERT on downstream tasks, BERT usually uses [CLS] token embeddings to represent a sentence, for example, in sentiment analysis tasks and [SEP] token if there are two sentences for inference tasks. In the BERT model, positional embeddings are added to token embeddings to help the model learn the positions of a sequence. Segment Embeddings are also used to differentiate two consecutive sentences by adding "0" to every token in first sentence sequence embeddings and "1" to second sentence sequence embeddings.

\subsection{Fine-tuned BERT Architectures}
The model's architecture takes the source tweet/target and its reply with [SEP] token between them. We duplicate source tweets to the number of replies for fine-tuning BERT.

We start with a pre-trained Arabic BERT model as a baseline, then apply the Average-Pooling layer to token embeddings to get a fixed-size vector and then concatenate it with [CLS] embeddings then forward it to the dense layer to fine-tune the model using cross-entropy loss function.

For BERT Models, we compare two methods: applying source tweet and reply to the model, and the second one is feeding target and reply to the model. For this task, we implemented three kinds of architecture to leverage BERT embeddings. The experiments with BERT models are as follows: 

\begin{figure}[htb]
	\centering
	\includegraphics[width=3.3in,height=2.8in]{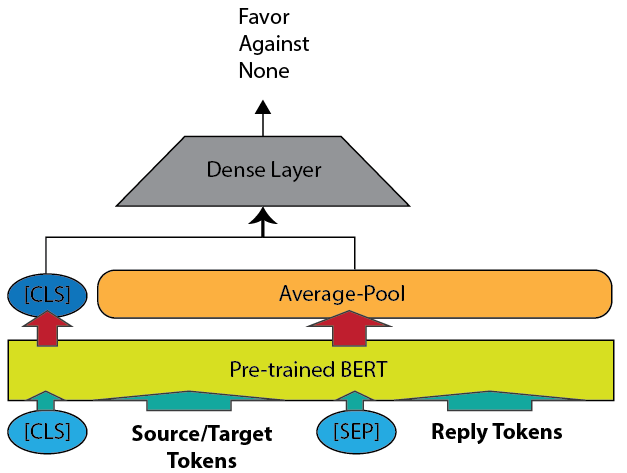}
	\caption{BERT Architecture with Average-Pool}
	\label{architecture}
\end{figure}
\begin{enumerate}
\item \textit{Fine-tune BERT using only [CLS] embeddings (BERT with sequence classification layer):} In this experiment whole source tweet or target is fed to the model alongside with reply and [SEP] token. In this method, the model learns to attend to the target in which replies are talking about it. Input to model is as follows: source/target + [SEP] + reply.
\item \textit{Fine-tune BERT with concatenation of [CLS] embeddings and average pooling layer on top of token embeddings:} In this experiment, the input of BERT is source/target + [SEP] + reply. The intuition for this method in addition to [CLS] embeddings, average pooling on token embeddings digests other tokens as shown in Figure \ref{architecture}.
\item \textit{Using frozen BERT embeddings for training with BiLSTM:} in this method, we used BERT as a feature extractor, so we averaged of last four layers of BERT token embeddings and fed them into Bidirectional LSTM and finally concatenate two hidden states of BiLSTM and fed them into the dense layer in the last. In this method source/ and reply are fed to BERT for extracting embeddings.
\end{enumerate}
The comparisons of mentioned models will be discussed in section \ref{experimental results} of the paper.

\section{Experimental Setups}\label{experimental setups}
In this section, we describe our model configurations and hyperparameters, and, in the last part, we discuss the evaluation metric used in the paper.
\subsection{Model Configuration and training}
For the base model, we experiment on \textit{bert-base-qarib\footnote{www.huggingface.co/qarib/bert-base-qarib}} \cite{ahmds21} model for fine-tuning our data. Because on Twitter Arabic users are from different countries and use different kinds of Dialects in the Arabic language, dialectical words could help to improve model performance. \textit{bert-base-qarib} was trained on 420M dialectical tweets and 180M texts sentences. The model includes 64k initial tokens, a dimension of 768 for hidden and embedding layers in addition to 12 Transformer encoder layers with 12 attention heads. Sentences used in this model without any tokenization but the mentions (@ + \textit{screen\_name}, e.g., \textit{@twitter}) and URLs deleted for source input and replies for the model. We used \textit{Huggingface\footnote{www.huggingface.co}} and \textit{PyTorch\footnote{ www.pytorch.org}} libraries for training and fine-tuning. 

\subsection{Hyperparameters tuning}
In hyperparameters selection for the BERT model, we experimented with Adam optimizer \cite{kb15} with weight decay of 0.01 and employing early stopping and with an initial learning rate of 3e-5 with default betas. We warmed up steps for training for about one epoch and also used a dropout rate of 0.4. All models trained on one Nvidia RTX 2070 GPU with 8GBs of ram and with a batch size of 16 and a maximum length of 128 for source tweet and reply sentences and fine-tuned for 5 epochs. To alleviate class imbalance between the None, positive and negative classes, we gave the Favor and Against classes twice as much weight as the None class.

For LSTM hyperparameters selection, we used Bidirectional LSTM and set the hidden size to 256. Optimizer parameters and other hyperparameters are like before.

\subsection{Evaluation}
We evaluate the performance of our data on about 2700 tweets with replies. We use \textit{F-score} for evaluation metric as a standard metric used in sentiment analysis and stance detection tasks. Semeval-2016 task 6 used macro averaged \textit{F-score} of two Favor and Against labels. The average of $F_{favor}$ and $F_{against}$ reported as an evaluation metric in this paper as the primary metric in the SemEval-2016 task. The formula of this metric is shown in Eq. \ref{favg}. Of which $F_{favor}$ and $F_{against}$ are calculated by Eq. \ref{ffavor} and Eq. \ref{fagainst} where $P_{favor}$, $P_{against}$, $R_{favor}$ and $R_{against}$ are precision and recalls of favor and against classes, respectively.

\begin{equation}\label{favg}
	F_{avg}=\dfrac{F_{favor}+F_{against}}{2}
\end{equation}

\begin{equation}\label{ffavor}
	F_{favor}=\dfrac{2P_{favor}R_{favor}}{P_{favor}+R_{favor}}
\end{equation}

\begin{equation}\label{fagainst}
	F_{against}=\dfrac{2P_{against}R_{against}}{P_{against}+R_{against}}
\end{equation}

\textit{F-scores} of None class were not disregarded; in information retrieval, it means None class is not of interest or negative class. F-score performs better than other metrics like accuracy in dominant classes because one dominant class can get high accuracy and misclassify every other instance.

\section{Experimental Results and Discussions}\label{experimental results}
Table \ref{results} presents different experimented methods results using Macro averaged F-scores of Favor, Against and Macro average of two classes described in the last section. We compare three methods using the BERT language model for the provided dataset. Each of them was tested on two different inputs besides reply: source tweet and target. For each model, we evaluate our test set on the model that has the highest validation Average F-score.  We highlight the best scores in bold.

Results show that target-only models perform a bit better than the whole source tweet sentence, and from this we can infer that, replies can attend well to target compared to target in the entire sentence. As future research, this can be investigated to represent the target better in the source tweet.

Results also show that [CLS] embeddings can digest information of two inputs (source/target + reply) very well. The best performing model on the dataset was \textit{Bert-seq-target}, described in the last section. This model used [CLS] embeddings only and got a maximum 70.69\% F-score. \textit{BERT-avgpool} uses both [CLS] embeddings and average pooling of token embeddings and, for the precision of negative class, performs a bit better than \textit{BERT-seq}. \textit{BERT-LSTM} performs the worst among other methods because BERT embeddings are frozen, so it could not leverage BERT pre-trained weights well enough, and it takes so much longer epochs to converge.

The most misclassified instances are between the None and Against classes. This is because of the nature of the discussion about the target in source tweets and replies. Replies can indirectly be against a target, but annotators could not find evidence in a reply to choose the Against option or vice versa. Because sometimes replies contain sarcasm, irony, or any other indirect talks toward a target and it is hard to find even for human annotators.

\section{Conclusion}
This paper provides a new general corpus for Target-based Stance detection with more than 200 targets in the Arabic Language. To the best of our knowledge, this is the first open-source Stance Detection corpus in Arabic. We also propose a new method to stance detection task using tweet replies to determine Stance toward a target in their source tweet. We compared different models and many different usages of representations from BERT for this task, fine-tuned them, and reached a 70.69\% F-Score. Consequently, our results proved the model could learn Stance detection with a wide variety of targets. Although the model performs poorly on some of the none and negative samples, the Maximum F-score of 70.69\% shows that it performs appropriately on a general dataset.

In future work, we aim to expand our dataset to have more targets and samples and experiment with other pre-trained models to improve the performance and F-Score. We can also study and test new methods on multi-target Stance detection and make the model learn the stances toward multiple targets in the source tweet.

\begin{table*}[tb]
		\renewcommand{\arraystretch}{1.5}
		\caption{Proposed Methods Results}
	\label{results}
	\centering
	\begin{threeparttable}
		\centering
	\begin{tabular}{|l|c|c|c|c|c|c|c|}
		\hline
		\textbf{Methods}             & \textbf{P\textsubscript{favor}} & \textbf{R\textsubscript{favor}} & \textbf{F\textsubscript{favor}} & \textbf{P\textsubscript{against}} & \textbf{R\textsubscript{against}} & \textbf{F\textsubscript{against}} & \textbf{F\textsubscript{avg}}  \\ \hline
		\textit{BERT-seq}            & 75.40           & \textbf{69.34}  & \textbf{72.24}  & 62.74             & 72.04             & 67.07             & 69.65          \\ \hline
		\textit{BERT-avgpool}        & 76.83           & 64.38           & 70.06           & 64.05             & 66.01             & 65.01             & 68.44          \\ \hline
		\textit{BERT-LSTM}           & 77.38           & 53.43           & 63.21           & 58.70             & 65.32             & 61.83             & 62.52          \\ \hline
		\textit{BERT-seq-target}     & \textbf{78.85}  & 64.23           & 70.80           & 62.63             & \textbf{80.83}    & 70.58             & \textbf{70.69} \\ \hline
		\textit{BERT-avgpool-target} & 71.69           & 63.94           & 67.59           & \textbf{64.18}    & 78.95             & \textbf{70.80}    & 69.20          \\ \hline
		\textit{BERT-LSTM-target}    & 71.70           & 59.56           & 65.07           & 63.94             & 71.15             & 67.35             & 66.21          \\ \hline
	\end{tabular}
\end{threeparttable}
\end{table*}

\ifCLASSOPTIONcompsoc
  \section*{Acknowledgments}
\else
  \section*{Acknowledgment}
\fi

The authors would like to thank \textit{Exa Company\footnote{www.exalab.co}} for providing the infrastructure, human resources, technical and non-technical knowledge which results to achieving this article and its goals.



%\bibliographystyle{IEEEtran}
%\begin{thebibliography}{1}
%\end{thebibliography}

\bibliographystyle{IEEEtran}
\bibliography{stancebiblio}

\end{document}